\documentclass[10pt,twocolumn,letterpaper]{article}

\usepackage[pagenumbers]{cvpr} 

\newcommand{\ours}{\textsc{Octopus}}
\usepackage{xcolor}
\usepackage{graphicx}
\usepackage{adjustbox}
\usepackage{adjustbox}
\usepackage{float} 
\usepackage{newtxtext}
\usepackage{algorithm}
\usepackage{amsmath}
\usepackage{algpseudocode}
\usepackage{marvosym}   

\usepackage{multirow}
\usepackage{makecell}      
\usepackage{booktabs}      
\usepackage{adjustbox}     
\usepackage{array}
\usepackage{colortbl}      
\usepackage{xcolor}        
\usepackage{graphicx}      
\usepackage{natbib}        
\usepackage{xcolor}
\definecolor{mydarkgreen}{rgb}{0.0, 0.5, 0.0} 
\newcommand{\better}[1]{\textcolor{ForestGreen}{\scriptsize{\,\,($\uparrow$#1)}}}

\newcommand{\betterrr}[1]{\textcolor{ForestGreen}{{+#1\%}}}

\usepackage{booktabs}
\usepackage{array}
\usepackage{xcolor}
\usepackage{adjustbox}
\usepackage{colortbl}   

\newcommand{\rcol}{\rowcolor{orange!12}}

\newcommand{\mypar}[1]{\vspace{5pt}\noindent\textbf{#1}~}
\usepackage{tcolorbox}
\newcommand{\mr}[1]{\mathrm{#1}}

\newcommand{\changes}[1]{{#1}}

\definecolor{cvprblue}{rgb}{0.21,0.49,0.74}
\usepackage[pagebackref,breaklinks,colorlinks,allcolors=cvprblue]{hyperref}
\usepackage{adjustbox}

\def\paperID{8832} 
\def\confName{CVPR}
\def\confYear{2026}

\title{
\textbf{
\ours{}: Enhancing the Spatial-Awareness of Vision SSMs with Multi-Dimensional Scans and Traversal Selection
}
}

\author{
\normalfont
Kunal Mahatha\textsuperscript{\Letter} \quad
Ali Bahri$^{1}$ \quad
Pierre Marza$^{2}$ \quad
Sahar Dastani$^{1}$ \quad
Maria Vakalopoulou$^{2}$ \\
Stergios Christodoulidis$^{2}$ \quad
Jose~Dolz$^{1}$ \quad
Christian~Desrosiers$^{1}$ \\[0.6em]
\small
$^{1}$ILLS, LIVIA, ÉTS Montréal, Canada \quad
$^{2}$MICS, CentraleSupélec, Université Paris-Saclay \\
\Letter\ \small \texttt{kunal.mahatha.1@ens.etsmtl.ca}\\[0.6em]
}

\begin{document}
\maketitle
\begin{abstract}

\noindent State space models (SSMs) have recently emerged as an alternative to transformers due to their unique ability of modeling global relationships in text with linear complexity. However, their success in vision tasks has been limited due to their causal formulation, which is suitable for sequential text but detrimental in the spatial domain where causality breaks the inherent spatial relationships among pixels or patches. As a result, standard SSMs fail to capture local spatial coherence, often linking non-adjacent patches while ignoring neighboring ones that are visually correlated. To address these limitations, we introduce \ours{}
, a novel architecture that preserves both global context and local spatial structure within images, while maintaining the linear complexity of SSMs. \ours{} performs discrete reoccurrence along eight principal orientations, going forward or backward in the horizontal, vertical, and diagonal directions, allowing effective information exchange across all spatially connected regions while maintaining independence among unrelated patches. This design enables multi-directional recurrence, capturing both global context and local spatial structure with SSM-level efficiency. In our classification and segmentation benchmarks, \ours{} demonstrates notable improvements in boundary preservation and region consistency, as evident from the segmentation results, while maintaining relatively better classification accuracy compared to existing V-SSM based models. These results suggest that \ours{} appears as a foundation method for multi-directional recurrence as a scalable and effective mechanism for building spatially aware and computationally efficient vision architectures. 
Our project page is publicly available at: \url{https://kunal-mahatha.github.io/octopus.page/}
\end{abstract}    
\section{Introduction}
\label{sec:intro}

\begin{figure}[t]
    \centering
    \includegraphics[width=.98\linewidth]{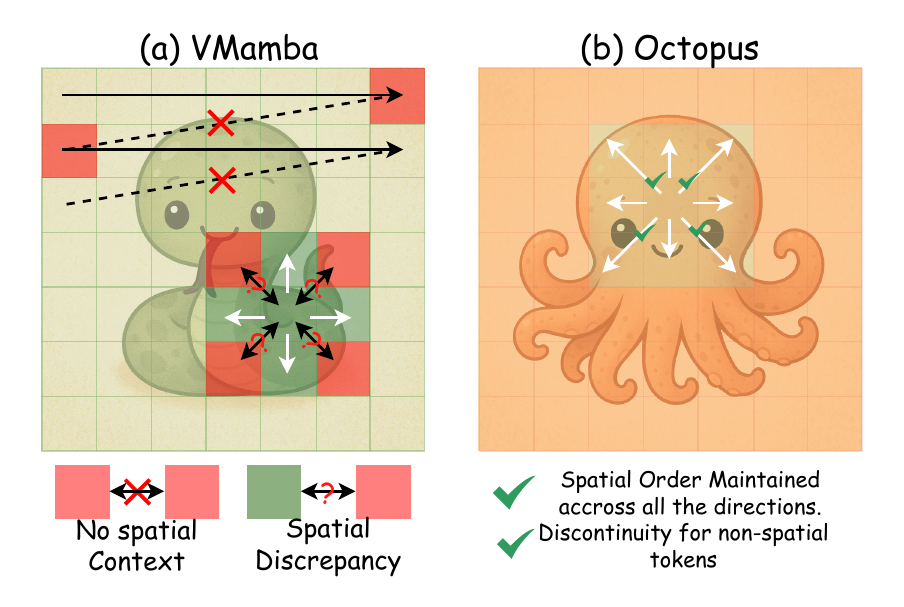}

    \vspace{-5pt}
    \caption{
    \textbf{Comparing directional scanning in VMamba and \ours{}.} 
    \textbf{(a) VMamba:} Uses continuous row-wise scans in four directions (\(\rightarrow, \leftarrow, \downarrow, \uparrow\)). 
    However, due to the continuity between rows, non-adjacent patches (e.g., the \textcolor{red}{red regions} across rows) become connected despite lacking spatial proximity — leading to \textit{no spatial context}. 
    Moreover, patches at actual spatial boundaries (from \textcolor{green!60!black}{green} to \textcolor{red}{red}) are not connected, creating a \textit{spatial discrepancy}. 
    \textbf{(b) \ours{}:} Employs discrete, independent scans across eight spatial directions (horizontal, vertical, and diagonal). 
    This preserves spatial order, ensuring that only spatially adjacent patches interact while maintaining complete directional recurrence across all neighborhoods.
    }
    \label{motovation}

    \vspace{-5pt}
\end{figure}

Understanding visual scenes requires more than recognizing individual objects: it demands grasping the spatial relationships and contextual continuity that define an image. From classification and detection to segmentation, modern computer vision models aim to encode such spatial structure into compact representations.

Among these tasks, segmentation remains the most spatially grounded, as it requires pixel-level reasoning and awareness of how neighboring regions interact to form coherent object boundaries.
The effectiveness of a model, therefore, depends not only on what features it learns, but on how information flows across spatial dimensions within the network.

The introduction of the Vision Transformer (ViT)~\cite{vaswani2017attention, doersch2015unsupervised} marked a major shift by leveraging self-attention to establish global dependencies between image patches.
This mechanism enabled long-range interactions and remarkable performance across a variety of vision tasks.
However, the quadratic complexity of self-attention limits its scalability to high-resolution images. Furthermore, its inherently dense interactions often lead to diffused attention patterns that struggle to maintain precise spatial alignment, a property crucial for segmentation and other dense prediction tasks~\cite{mahatha2025nerveneighbourhoodentropyguided}.

To address these limitation of quadratic complexity, State-Space Models (SSMs)~\cite{mamba, mamba2} have recently emerged as a compelling alternative.
Originally developed for sequential data, SSMs describe how information evolves over time through recurrent state transitions, offering linear computational complexity and efficient memory usage in practice.
Inspired by their success in language modeling, several works have adapted SSMs for visual understanding, resulting in the rapidly emergent family of Vision SSMs (VSSMs)~\cite{liu2024vmamba, huang2024localmamba, xie2024quadmamba, zhu2024vision, hatamizadeh2024mambavision, Bahri_2025_CVPR}.
These models reinterpret an image as a sequence of tokens scanned in raster order, applying recurrent updates across spatial positions to model long-range dependencies efficiently.

While effective in achieving global context with linear efficiency, this one-dimensional sequential formulation invokes causality which is meaningful in text, where order conveys semantics, but in vision introduces it a fundamental limitation.
By scanning tokens along a fixed order, typically from left to right and top to bottom, existing VSSMs inherit a directional bias that conflicts with the inherently two-dimensional nature of visual data.
As a result, information tends to propagate predominantly along one or two axes, weakening spatial coherence—especially across vertical and diagonal neighborhoods.
\textit{As illustrated in Fig.~\ref{motovation}}, a continuous raster scan (as used in VMamba) inadvertently links patches \emph{(sequential modeling)} that are spatially disconnected (e.g., red–red transitions across rows), while simultaneously failing to establish direct interactions between diagonally adjacent patches that are physically close \emph{(spatial geometry)} in the image causing a spatial discrepency.
This mismatch between the scan trajectory and the true 2D neighborhood structure underscores the need for a directionally balanced formulation such as \ours{}.

This misalignment between \emph{sequential modeling} and \emph{spatial geometry} motivates a deeper rethinking of how state-space dynamics can be adapted to visual understanding.
If vision fundamentally depends on how information propagates through space, then we must ask whether our models can truly maintain uniform spatial continuity across all directions.
This line of reasoning leads naturally to a simple yet revealing thought experiment, which forms the foundation of our proposed \ours{} framework.

\textbf{Question 1:} 
\textit{Can the model propagate information equally well in all spatial directions: horizontal, vertical, and diagonal?} 

\textbf{Question 2:}
\textit{Can it ensure
isotropic and spatially coherent representations?}

\noindent These are the central questions that motivates our work. 
We argue that an isotropic model, one that treats all spatial directions equally, is essential for capturing the rich 2D relationships in visual data. 

However, simply increasing the number of scan directions does not inherently fix these issues.
\emph{Prior analyses of VMamba}~\cite{scanvmamba} show that naively extending continuous scans to eight directions does not improve the performance, in many cased it can even tend to degrade, because the underlying recurrence still follows a single elongated traversal rather than respecting true 2D neighborhoods.
Moreover, multi-directional recurrences introduce additional complications: multiple traversals tend to amplify the effective transition matrix dynamics (“diagonal blow-up”), and aggregating multiple bi-directional scans becomes non-trivial—while aggregation is manageable with one or two directions, combining many directional streams risks overwhelming or misalignment spatial cues.
These observations indicate that expanding directional capacity must be done with considerable caution, as both the recurrence dynamics and the aggregation strategy fundamentally shape spatial coherence.

To address these challenges, we introduce \ours{}, a spatial-aware VSSM that achieves isotropic information propagation through discrete multi-directional scans.
By scanning features along eight principal directions and merging them through a spatially-consistent fusion mechanism, \ours{} restores spatial coherence while preserving the linear complexity of state-space models.
This design enables uniform feature propagation across all spatial dimensions, maintaining continuity among neighboring patches while suppressing irrelevant interactions between non-adjacent regions. By aligning state-space dynamics with the inherent two-dimensional geometry of images, \ours{} achieves spatial awareness, and linear efficiency, advancing the capability of state-space models for vision understanding.

\noindent Our main contributions can be summarized as follows:
\begin{itemize}
    \item We propose \ours{}, an 8-directional spatial-aware VSSM enabling isotropic feature propagation and spatial continuity. The key innovation of \ours{} is to efficiently capture the complete \emph{spatial relationships among image patches while suppressing information flow between non-adjacent regions}, ensuring strong spatial coherence.
    \item We introduce a \emph{discrete scan} formulation that performs independent state updates along eight principal directions instead of a continuous raster scan. To prevent matrix explosion from directional accumulation, we introduce the \emph{Normalized Transition Kernel}. Furthermore, \emph{a traversal selection mechanism} is proposed to enable careful directional fusion, preserving spatial structure and consistency.

    \item Through our experiments on image classification and segmentation benchmarks, we demonstrate that \ours{} attains state-of-the-art performance in segmentation with strong spatial consistency, while maintaining consistent and better classification accuracy compared with previous V-SSM methods, while preserving the linear computational and memory efficiency inherent to state-space models.

\end{itemize}
\section{Related Works}
\label{sec:related_works}

\begin{figure*}[t]
    \centering
    \includegraphics[width=1.0\textwidth]{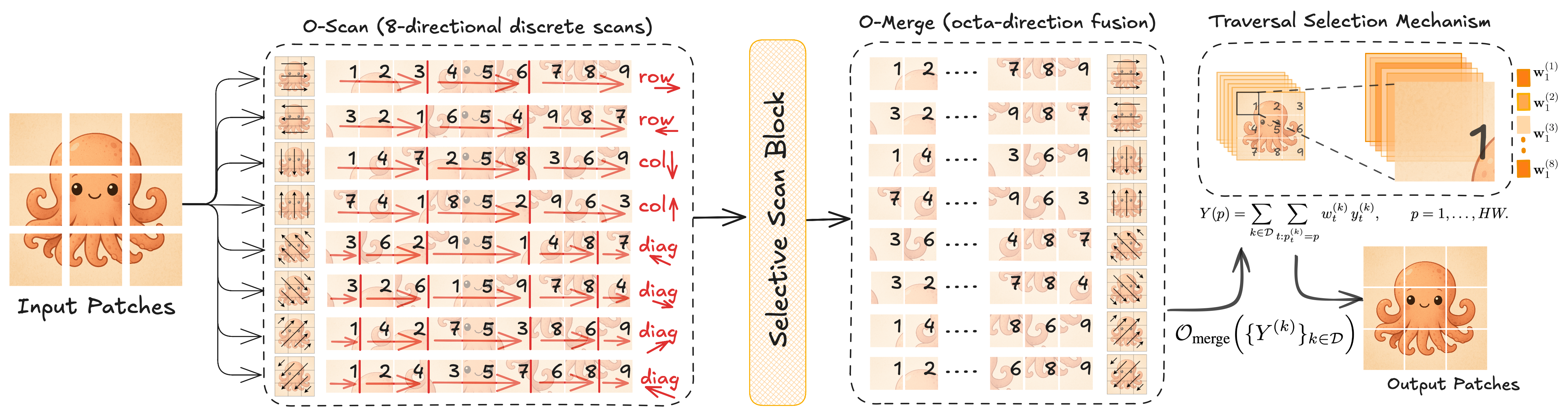}
\caption{\textbf{Overview of our pipeline.} O-Scan performs eight discrete canonical traversals,
$\{\mr{row}_{\!\scriptscriptstyle\rightarrow},\;
\mr{row}_{\!\scriptscriptstyle\leftarrow},\;
\mr{col}_{\!\scriptscriptstyle\downarrow},\;
\mr{col}_{\!\scriptscriptstyle\uparrow},\;
\mr{diag}_{\!\scriptscriptstyle\searrow},\;
\mr{diag}_{\!\scriptscriptstyle\nwarrow},\;
\mr{diag}_{\!\scriptscriptstyle\swarrow},\;
\mr{diag}_{\!\scriptscriptstyle\nearrow}\}$,
covering the complete set of horizontal, vertical, and diagonal orientations. 
The \textcolor{red}{\textbf{\texttt{|}}} markers in O-Scan denote boundaries between independent scan-lines, where the hidden state is reset, forming the core of our \textit{discrete scan} formulation. 
Each directional sequence is processed by the \textit{Selective Scan Block}, fused via \textit{O-Merge}, and aggregated with the \textit{Traversal Selection Mechanism} to produce the final output patches.
}
    \label{fig:pipeline}

    \vspace{-5pt}
\end{figure*}

\mypar{State Space Models (SSMs).}  State-space models (SSMs) have emerged as a powerful alternative to Transformers for long-sequence processing \cite{gu2021combining, gu2021efficiently, smith2022simplified, mehta2022long}. Among these, the Structured State Space Sequence (S4) model \cite{gu2021efficiently} stands out for its efficient treatment of long-range dependencies through a diagonal parameterization, which mitigates the computational bottlenecks of earlier approaches. Recent progress in sequence modeling has been propelled by the development of parallel scanning methods, inspired by models like S4, S5 \cite{smith2022simplified}, and H3 \cite{fu2022hungry}. Building on this line of work, Mamba \cite{gu2023Mamba} extends S4 by introducing a data-dependent selection mechanism, enabling linear scalability. This innovation allows Mamba to handle longer sequences efficiently while maintaining strong performance and avoiding exponential growth in computational requirements. Such advances are particularly valuable for large-scale or real-time applications, including natural language processing (NLP) and time-series forecasting.

\mypar{SSMs for visual understanding.} The success of Mamba models in NLP has sparked substantial advancements in computer vision, inspiring several adaptations designed specifically for visual data. Researchers have proposed multiple extensions of the Mamba architecture to tackle vision-specific challenges. For example, VMamba \cite{liu2024vmamba} introduces an SS2D module to handle direction sensitivity that arises between non-causal 2D images and sequential 1D data. LocalMamba \cite{huang2024localmamba} emphasizes localized feature modeling by employing state-space layers to strengthen neighborhood interactions while reducing reliance on self-attention, thus enabling efficient analysis of high-resolution images. Meanwhile, QuadMamba \cite{xie2024quadmamba} adopts a quadrilateral traversal strategy to capture directional dependencies within 2D images, effectively integrating information from different image regions.

\mypar{2-dimensional SSMs.} Methods that convert 2D images into 1D sequences by scanning them along different directions face a spatial discrepancy problem: consecutive tokens in the sequence may represent image regions that are not spatially related, disrupting the image’s structural coherence. To overcome this, 2D SSMs preserve the spatial continuity inherent to two-dimensional data such as images \cite{baron20242,fillioux2023structured}, though their sequential processing makes them computationally slow. To enhance efficiency, 2DMamba \cite{zhang20252dmamba} introduced a two-stage strategy: image patches are first processed row by row in parallel, and the resulting hidden states are then used to perform parallel vertical scans at corresponding positions. By optimizing this 2D selective scanning process for hardware execution, 2DMamba achieves improvements in both accuracy and computational speed for whole-slide image (WSI) classification.

Our work differs from VMamba, which flattens images into long raster sequences with spatial discontinuity, and from 2DMamba, whose simple two-step row-column scanning strategy does not support bidirectional and diagonal scanning contexts, limiting downstream tasks like segmentation. In contrast, we introduce discrete multi-line scans along eight canonical orientations and use a normalized transition kernel together with a traversal-selection mechanism to fuse directional evidence, enabling isotropic and spatially coherent state-space propagation.

\section{Methodology}
\label{sec:method}

\subsection{Preliminaries}

\mypar{State Space Models (SSMs).} 
Originating from the Kalman filter, SSMs are \emph{linear time-invariant (LTI)} systems that map an input sequence $x(t)$ to an output $y(t)$ through a hidden state $h(t) \in \mathbb{R}^N$:
\begin{equation}
    \frac{d}{dt}h(t) = A\,h(t) + B\,x(t), \qquad y(t) = C\,h(t),
\label{eq:ssm_continuous_short}
\end{equation}
where $A,B,C$ govern the state transition, input projection, and readout mappings.

\mypar{Discretization.}
Using the Zero-Order Hold (ZOH) rule with step $\Delta$, the system becomes
\begin{equation}
    h_t = \bar{A}\,h_{t-1} + \bar{B}\,x_t, \qquad y_t = C\,h_t,
\label{eq:ssm_discrete_short}
\end{equation}
with discretized parameters 
$\bar{A}\!=\!e^{\Delta A}$ and 
$\bar{B}\!=\!(\Delta A)^{-1}(e^{\Delta A}-I)B \approx \Delta B$.

\mypar{Selective SSMs (Mamba).}
Mamba extends SSMs with a learned selective gate $f_t$ controlling dynamic recurrence:
\begin{equation}
    h_t = f_t \odot (\bar{A}h_{t-1} + \bar{B}x_t), 
    \qquad y_t = C\,h_t.
\label{eq:mamba_short}
\end{equation}
This formulation enables long-range, content-adaptive modeling with \emph{linear complexity}.
\ours{} builds on this foundation by extending the selective SSM to \emph{eight spatial directions} 
for structured, isotropic propagation across the image plane.

\subsection{OCTOPUS}

\subsubsection{O-Scan}
\label{subsubsec:o_scan}

Let $X \in \mathbb{R}^{B\times C\times H\times W}$ denote a spatial feature map, where $B$ is the batch size, $C$ the channel dimension, and $H\times W$ the spatial resolution.
Existing vision state-space models linearize $X$ into a single raster sequence and apply a one-dimensional recurrence of the form
\begin{equation}
    h_t = A h_{t-1} + B x_t, \quad y_t = C h_t,
\end{equation}
which models dependencies along a fixed scan order.  
Although efficient, this formulation constrains information flow to a single axis and therefore introduces \textit{directional bias}, propagating features unevenly across the spatial plane.

\mypar{Directional Decomposition.}
To enable isotropic feature propagation, we define a discrete set of eight canonical orientations shown in Figure~\ref{fig:pipeline}:
\begin{equation}
\begin{aligned}
\mathcal{D} = 
\{&\mr{row}_{\!\scriptscriptstyle\rightarrow}, \mr{row}_{\!\scriptscriptstyle\leftarrow},
\mr{col}_{\scriptscriptstyle\downarrow}, \mr{col}_{\scriptscriptstyle\uparrow},\\
&\mr{diag}_{\scriptscriptstyle\searrow}, \mr{diag}_{\scriptscriptstyle\nwarrow},
\mr{diag}_{\scriptscriptstyle\swarrow},
\mr{diag}_{\scriptscriptstyle\nearrow}\}.
\end{aligned}
\end{equation}

Each direction $k \in \mathcal{D}$ (e.g., every row, every column, or every diagonal) consists of \emph{multiple discrete scan-lines}.  
Each scan-line $\ell$ induces an independent traversal order
\begin{equation}
    \pi^{(k,\ell)} = \big(p^{(k,\ell)}_1,\, p^{(k,\ell)}_2,\, \dots,\, p^{(k,\ell)}_{L_{k,\ell}}\big),
\end{equation}
where $p^{(k,\ell)}_t = (i^{(k,\ell)}_t,\, j^{(k,\ell)}_t)$ denotes the spatial coordinates visited at step $t$.  
The hidden state is reset between scan-lines, ensuring that information propagates only along itself which is aligned with orientation $k$, rather than forming a single long 1D sequence.

Given all scan-lines, the \emph{O-Scan} operator rearranges the spatial features of $X$ into directional sequences:
\begin{equation}
    X^{(k,\ell)} = \big\{X(p^{(k,\ell)}_t)\;\big|\; t = 1,\dots, L_{k,\ell}\big\},
    \qquad k \in \mathcal{D}.
\end{equation}
Each $X^{(k,\ell)}$ represents a discrete 1D projection of the feature map along orientation $k$, analogous to “reading’’ the image in that direction.

\mypar{Directional Recurrence.}
State propagation is then applied independently along each scan-line of each orientation:
\begin{equation}
    h^{(k)}_t = A^{(k)} h^{(k)}_{t-1} + B^{(k)} x^{(k)}_t, 
    \quad y^{(k)}_t = C^{(k)} h^{(k)}_t,
\end{equation}
where $A^{(k)}, B^{(k)}, C^{(k)}$ are learnable transition matrices shared across steps in direction $k$.  
This formulation enables the model to capture context along multiple spatial paths while maintaining the linear complexity of the standard state-space formulation.

\mypar{Spatial Interpretation of the O-Scan.}
By scanning features along eight complementary directions, \emph{O-Scan} transforms a single anisotropic propagation into an \textit{octa-directional spatial process}.  
Each direction captures dependencies consistent with its geometric path (horizontal, vertical, or diagonal), while their union covers the full image topology.  
This design ensures that information is propagated uniformly across orientations, promoting \textit{spatial continuity} among neighboring regions and suppressing unintended flow between spatially disjoint areas.  
As a result, \emph{O-Scan} establishes the foundation for \textit{isotropic feature propagation}, bridging the gap between one-dimensional recurrence and two-dimensional visual geometry.

\begin{figure*}[t]
    \centering
    \includegraphics[width=1.0\textwidth]{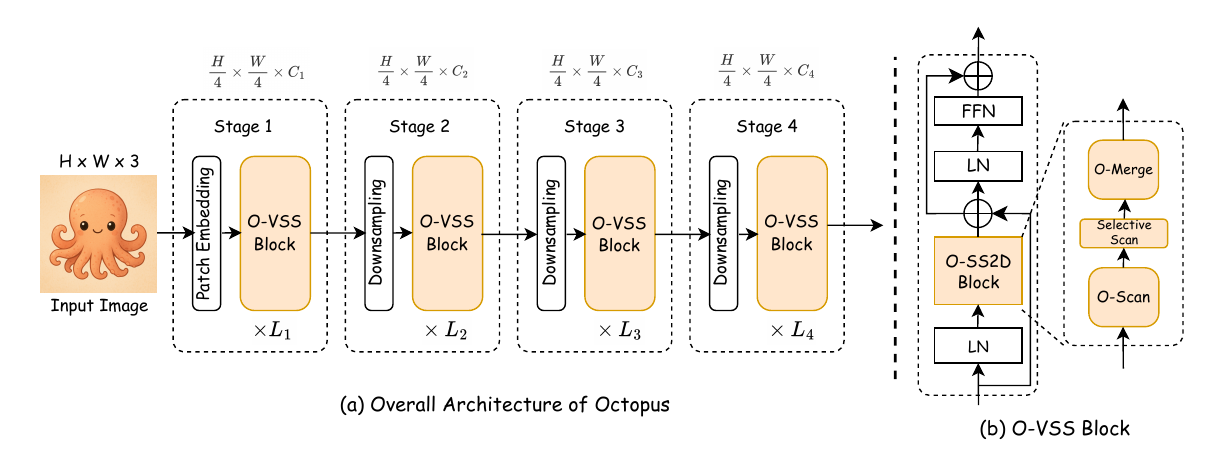}

    \vspace{-5pt}
    \caption{\textbf{Overall Architecture of Octopus:} (a) The full hierarchical encoder consisting of four stages. Each stage applies multiple O-VSS blocks, with spatial resolution preserved inside the stage and reduced via downsampling between stages. (b) The O-VSS block design, composed of Layer Normalization, an O-SS2D block (which integrates the proposed 8-direction Selective Scan), and an FFN. The O-SS2D block itself combines O-Scan operations with an O-Merge module to fuse multi-directional features.}
    \label{fig:arch}

    \vspace{-5pt}
\end{figure*}

\subsubsection{State Propagation and Structured Normalization}

\mypar{Structured Matrix View.}
Each directional recurrence corresponds to multiplication by a semiseparable matrix:
\begin{equation}
y = \mathrm{SSM}(A_k,B_k,C_k)(x) = M_k x, 
\\
(M_k)_{ji} = C_k^\top A_k^{\,j-i} B_k,
\label{eq:ssm_matrix}
\end{equation}
where $M_k \in \mathbb{R}^{T\times T}$ has lower-triangular blocks of rank at most $d_{\text{state}}$.  
This formulation preserves the linear-time complexity of SSMs along each directional sequence.

\mypar{Normalized Transition Kernel.}
Directly summing the directional matrices,
\(
M = \sum_{k} M_k,
\)
over-amplifies diagonal terms because each $M_k$ contributes the identity $A_k^0 = I$.  
This leads to \emph{diagonal dominance}, causing self-recurrence to overshadow spatial propagation and breaking isotropy.

To maintain balanced dynamics, \ours{} adopts a \emph{normalized kernel}:
\begin{equation}
A_k = \tfrac{A}{8}, \qquad B_k = B, \qquad C_k = C, \qquad k \in \mathcal{D}.
\label{eq:ak_normalized}
\end{equation}
This ensures the equilibrium condition $\sum_{k} A_k = A$, preserving the overall transition strength of a single SSM while distributing it evenly across all orientations.

Under this normalization, each $M_k$ has entries 
$(M_k)_{ji} = C^\top (A/8)^{\,j-i} B$, 
so the aggregated matrix satisfies
\begin{equation}
M = \sum_{k=1}^8 M_k,
\end{equation}
where the diagonal and off-diagonal terms scale proportionally.  
This prevents the effective recurrence coefficient from growing to $8|a|$ and preserves the stability condition $|a|<1$.

\mypar{Isotropic Structured Dynamics.}
The normalized kernel yields an \emph{isotropic} transformation matrix $M$: 
each directional SSM contributes equally, preventing any single orientation from dominating the propagation.  
As a result, $M$ remains balanced and semiseparable, enabling uniform information diffusion across all eight principal directions and supporting stable, geometry-aware state propagation within \ours{}.

\subsubsection{O-Merge: Octa-Directional Fusion with Traversal Selection}
\label{subsubsec:o_merge}

Following the directional propagation in \emph{O-Scan}, each orientation 
$k \in \mathcal{D}$ produces multiple output sequences
\[
Y^{(k,\ell)} = \{ y^{(k,\ell)}_1,\, y^{(k,\ell)}_2,\, \dots,\, y^{(k,\ell)}_{L_{k,\ell}} \},
\]
one for each discrete scan-line $\ell$ aligned with direction $k$.  
These sequences encode spatial dependencies along their respective 
one-dimensional traversal paths.  
\emph{O-Merge} reconstructs the global feature map by mapping all 
directional outputs back onto the two-dimensional grid and fusing them in 
a data-dependent manner.

\mypar{Inverse Mapping.}
Each scan-line defines an inverse placement operator
\[
\mathcal{S}^{(k,\ell)^{-1}} : \mathbb{R}^{L_{k,\ell}} \rightarrow \mathbb{R}^{H\times W},
\]
which restores the spatial layout by placing each token at its original 
spatial coordinate:
\[
\big[\mathcal{S}^{(k,\ell)^{-1}}(Y^{(k,\ell)}) \big]_{p^{(k,\ell)}_t}
    = y^{(k,\ell)}_t.
\]
Because multiple directional scan-lines intersect at many spatial 
locations, their contributions must be aggregated during reconstruction.

\mypar{Traversal Selection Mechanism (O-Attention).}
To avoid treating all directions equally, and inspired by the 
squeeze-and-excitation formulation~\cite{hu2018squeeze}, we introduce a 
lightweight selection module, \emph{O-Attention}, which learns 
patch-wise directional relevance.  
After inverse mapping, we stack all directional features at each spatial 
location:
\[
X(p) \in \mathbb{R}^{C \times 8},
\qquad p = 1,\dots, HW,
\]
where the eight channels correspond to the eight orientations.

O-Attention applies two pointwise (1$\times$1) Conv1D projections:
\[
u^{(k)}(p) = \phi_2(\phi_1(X^{(k)}(p))),
\]
and normalizes them across directions:
\[
w^{(k)}(p) = 
\frac{\exp\big(u^{(k)}(p)\big)}
     {\sum_{k' \in \mathcal{D}} \exp\big(u^{(k')}(p)\big)} ,
\qquad
k \in \mathcal{D}.
\]
This yields a spatially varying probability distribution over traversal 
directions.

\mypar{Spatial Fusion.}
Directional contributions are fused according to their learned weights:
\begin{equation}
\label{eq:o-merge-attn-corrected}
Y(p)
= \sum_{k \in \mathcal{D}}
  \sum_{\ell}
  \sum_{t: p^{(k,\ell)}_t = p}\!\!\!
  w^{(k)}(p)\, y^{(k,\ell)}_t ,
\quad p = 1,\dots, HW.
\end{equation}
This corresponds to a masked scatter-add across all scan-lines and 
all directions.  
Finally, the O-Merge operator can be written as:
\[
\mathcal{O}_{\text{merge}}
\Big( \{Y^{(k,\ell)}\}_{k \in \mathcal{D}, \ell} \Big)
=
\sum_{k \in \mathcal{D}}
\sum_{\ell}
\mathcal{S}^{(k,\ell)^{-1}}
\!\big( W^{(k)} \odot Y^{(k,\ell)} \big),
\]
where $W^{(k)} = w^{(k)}(p)$ is broadcast across channels.

\mypar{Interpretation.}
\emph{O-Merge} fuses the outputs of all scan-lines across eight 
complementary traversal orientations into a coherent spatial representation.  
The integrated \emph{O-Attention} module performs a 
\emph{traversal selection mechanism}, amplifying orientations that match 
local geometry while suppressing conflicting or less informative flows.  
This yields an orientation-balanced, spatially isotropic reconstruction 
while preserving the linear-time efficiency of the underlying 
directional state-space updates. This unified representation serves as the output of the \ours{} block, 
ready for downstream tasks such as segmentation or recognition.
\section{Experiments}
\label{sec:experiment}

\mypar{Datasets.} Following the recent SpectralMamba \cite{dastani2025spectral}, we resort to \textit{mini}ImageNet \cite{imagenet15russakovsky} for classification, and to Ade20K \cite{zhou2017scene} for segmentation. In particular, \textit{mini}ImageNet includes 50,000 training images and 10,000 validation images across 100 categories. ADE20K includes 150 fine-grained semantic categories and comprises 20,000 training images, 2,000 validation images, and 3,000 test images. 

\mypar{Baselines.} We evaluate \ours{} against strong baselines in both classification and semantic segmentation. For classification, 
we compare our approach with \textit{transformer-based} models (DeiT~\cite{DeiT2021}, Swin~\cite{Swin2021}, XCiT~\cite{xcit}) and \textit{state-space} models, including Vim, LocalViM~\cite{zhu2024vision}, MSVMamba~\cite{shi2024multi}, and VMamba~\cite{liu2024vmamba} across comparable tiny and small scales. For segmentation, 
we benchmark \ours{} against VMamba \cite{liu2024vmamba} and the stronger Spectral VMamba \cite{dastani2025spectral} variants under both single-scale and multi-scale testing.

\mypar{Architecture details.} An overview of \ours{} is shown in Figure~\ref{fig:arch}(a), with full 
configurations provided in the Appendix. The input image is tokenized by a 
patch embedding layer, producing a feature map of size $H/4 \times W/4$. The 
hierarchical encoder then processes features across four stages with 
resolutions $H/4$, $H/8$, $H/16$, and $H/32$, where each stage (except the 
first) begins with a downsampling layer followed by a stack of O-VSS blocks. The O-VSS blocks, illustrated in Figure~\ref{fig:arch}(b), serve as the visual 
counterparts to VSS/Mamba-style modules. Each block incorporates our O-SS2D 
module (Figure~\ref{fig:arch}(c)), which combines an 8-direction discrete 
O-Scan with an O-Merge fusion mechanism. O-Scan performs independent 
horizontal, vertical, and diagonal traversals with hidden-state resets, while 
O-Merge reconstructs the spatial layout and selectively aggregates directional 
features through the traversal-selection mechanism. This design enables 
isotropic spatial propagation while retaining the linear computational 
efficiency of state-space models. All OCTOPUS models in this paper are built 
using this improved O-VSS structure.

\mypar{Implementation details.} \textit{Image Classification:} Following previous works \cite{Swin2021,liu2024vmamba,zhu2024vision}, we train our models for 300 epochs with a batch size of 64 per GPU. The models are optimized using the AdamW optimizer with a momentum of 0.9. A cosine decay scheduler manages the learning rate, which starts at $5 \times 10^{-4}$, alongside a weight decay of 0.05. Additionally, we apply an exponential moving average (EMA) to stabilize training. For input images of size $224 \times 224$, our data augmentation techniques include color jittering, AutoAugment, random erasing, mixup, and cutmix. \textit{Image Segmentation:} we fine-tuned the model similar to\cite{zhu2024vision, ng2001spectral}, originally pre-trained on the \textit{mini}ImageNet dataset—to perform segmentation tasks on ADE20K dataset \cite{zhou2019semantic}. 
We use AdamW with a weight decay of 0.01 and a total batch size of 2 per GPU. The training schedule features an initial learning rate of 6\,$\times$\,10$^{-5}$, linear decay, a 1500-iteration linear warmup, and a total of 160,000 iterations. We apply standard data augmentations such as random horizontal flipping, random scaling within a ratio range of 0.5 to 2.0, and random photometric distortion.

\subsection{Main Results}

\mypar{Image Classification.}

Across both Tiny and Small scales, \ours{} consistently outperforms existing SSM-based models while remaining competitive with its Transformer-based counterparts. \ours{}-T achieves a Top-1 accuracy of 86.60\%, surpassing VMamba-T by \betterrr{\textbf{0.78}}, and also improves Top-5 accuracy to 96.63\% \betterrr{\textbf{0.26}}. Similarly, \ours{}-S reaches 86.89\% Top-1 accuracy, yielding a +0.41\% gain over VMamba-S, along with a Top-5 accuracy of 96.81\%. These gains demonstrate that the proposed multi-directional discrete scanning and traversal-selection fusion enable stronger spatial reasoning compared to prior VSSMs, leading to more discriminative visual representations without increasing computational cost excessively.

\begin{table}[t!]
\centering
\footnotesize
\setlength{\tabcolsep}{8pt}
\renewcommand\arraystretch{1.05}
\caption{\textbf{Classification performance on \textit{mini}ImageNet.} 
Input images are of size $224 \times 224$. T, S, B, M and N denote the tiny, small, base, micro and nano scales across each method.}
\label{main_table}
\resizebox{\columnwidth}{!}{
\begin{tabular}{l|c|cc}
\toprule
\textbf{Model} & \textbf{FLOPs (G)} & \textbf{Top-1 (\%)} & \textbf{Top-5 (\%)} \\
\midrule

\multicolumn{4}{c}{\textbf{Transformer-Based}} \\
\midrule
DeiT-S~\cite{DeiT2021} & 4.6 & \changes{70.83} & \changes{89.74} \\
DeiT-B~\cite{DeiT2021} & 17.5 & \changes{72.43} & \changes{90.14} \\
Swin-T~\cite{Swin2021} & 4.5 & 83.25 & 95.54 \\
Swin-S~\cite{Swin2021} & 8.7 & 84.10 & 95.53 \\
Swin-B~\cite{Swin2021} & 15.4 & 82.77 & 95.33 \\
XCiT-S24~\cite{xcit}   & 9.2 & 85.79 & 96.31 \\
XCiT-M24~\cite{xcit}   & 16.2 & 86.80 & 96.38 \\
\midrule

\multicolumn{4}{c}{\textbf{SSM-Based}} \\
\midrule
Vim-T~\cite{zhu2024vision}      & 1.5 & 67.30 & 87.97 \\
Vim-S~\cite{zhu2024vision}      & 5.1 & 79.70 & 93.20 \\
LocalVim-T~\cite{huang2024localmamba} & 1.5 & 82.12 & 94.60 \\
LocalVim-S~\cite{huang2024localmamba} & 4.8 & 81.68 & 93.63 \\
MSVMamba-N~\cite{shi2024multi} & 0.9 & 82.16 & 95.10 \\
MSVMamba-M~\cite{shi2024multi} & 1.5 & 83.72 & 95.81 \\
MSVMamba-T~\cite{shi2024multi} & 4.6 & 86.48 & 96.43 \\
VMamba-T~\cite{liu2024vmamba}  & 4.9 & 85.82 & 96.37 \\
VMamba-S~\cite{liu2024vmamba}  & 8.7 & 86.48 & 96.79 \\
\midrule

\rcol
\ours{}-T & 6.1 & 86.60 \textbf{\better{0.78}} & 96.63 \textbf{\better{0.26}} \\
\rcol
\ours{}-S & 10.0 & 86.89 \textbf{\better{0.41}} & 96.81 \textbf{\better{0.02}} \\

\bottomrule
\end{tabular}
}
\end{table}

\noindent \textbf{Image Segmentation.}
Table~\ref{tab:segmentation} reports the segmentation performance on ADE20K under both single-scale (SS) and multi-scale (MS) evaluation. Across all settings, \emph{\ours{}-T} significantly outperforms the different VMamba and SpectralVMamba baselines. More concretely, in the single-scale testing, \ours{}-T achieves an mIoU of 37.93, providing a substantial improvement of \betterrr{\textbf{15.16}} over VMamba-T and \betterrr{\textbf{11.41}} over the highest-performing SpectralVMamba variant, i.e., SpectralVMamba-B. A similar trend is observed in the multi-scale evaluation, where \ours{}-T reaches an mIoU of \emph{37.98}, outperforming prior VSSM-based models by wide margins. These improvements underscore the effectiveness of \ours{}'s discrete directional scanning and spatial aggregation mechanisms, which enhance spatial reasoning without necessarily increasing model size. As reflected in the results, this is particularly important in dense prediction tasks, such as segmentation, where fine-grained spatial understanding and efficient representation are critical for handling complex spatial configurations and identifying object boundaries.

\begin{table}[h!]
\centering
\footnotesize
\setlength{\tabcolsep}{12pt}
\caption{\textbf{Results of semantic segmentation on ADE20K.} 
SS and MS denote single-scale and multi-scale testing, respectively.}
\label{tab:segmentation}
\vspace{-3pt}

\resizebox{\columnwidth}{!}{
\begin{tabular}{l|cc}
\toprule
\textbf{Method}    & \textbf{mIoU (SS)} & \textbf{mIoU (MS)} \\ 
\midrule
VMamba-T~\cite{liu2024vmamba}            & 22.77 & 23.98 \\ 
VMamba-S~\cite{liu2024vmamba}            & 25.84 & 27.13 \\ 
VMamba-B~\cite{liu2024vmamba}            & 26.32 & 28.41 \\ 
SpectralVMamba-T~\cite{dastani2025spectral}    & 25.58 & 27.72 \\ 
SpectralVMamba-S~\cite{dastani2025spectral}    & 27.52 & 29.34 \\ 
SpectralVMamba-B~\cite{dastani2025spectral}    & 27.81 & 28.99 \\ 
\midrule
\rcol
\ours{}-T            & 37.93 \textbf{\better{15.16}} & 37.98 \textbf{\better{14.00}} \\ 
\rcol
\ours{}-S            & 39.04 \textbf{\better{13.12}} & 39.32 \textbf{\better{12.19}} \\ 
\rcol
\ours{}-B            & 40.32 \textbf{\better{14.00}} & 40.95 \textbf{\better{12.54}} \\ 
\bottomrule
\end{tabular}
}
\vspace{0.5pt}
\end{table}

\subsection{Ablation Studies and Qualitative Analysis}
\begin{figure}[h!]
    \centering
    \includegraphics[width=0.96\columnwidth]{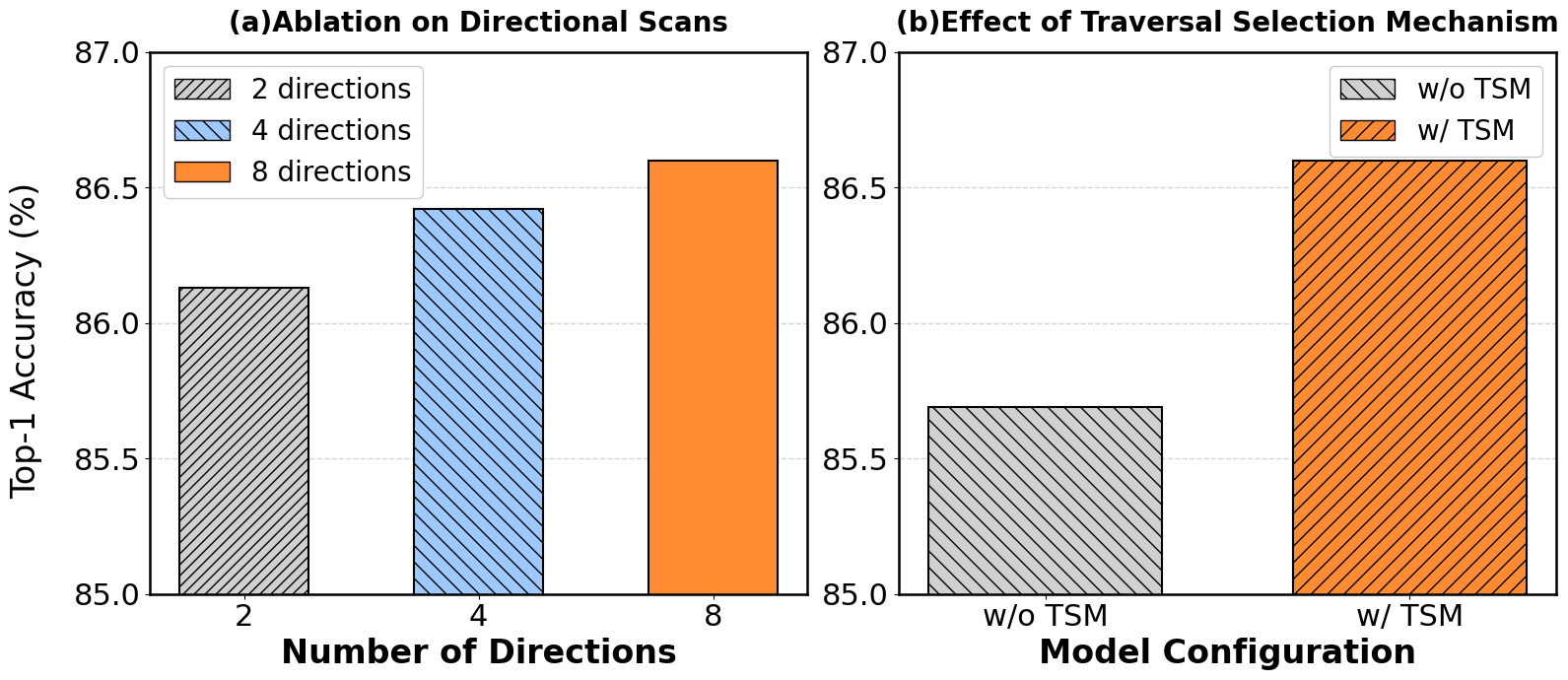}

    \vspace{-5pt}
    \caption{\textbf{Ablation on directional scans and traversal selection.} 
    \textbf{Left:} Increasing the number of canonical scan directions from 2 to 8 steadily improves recognition performance, validating the effectiveness of our discrete multi-line scan formulation. 
    \textbf{Right:} Incorporating the Traversal Selection Mechanism (TSM) further enhances Top-1 accuracy by enabling data-dependent fusion of directional evidence.}
    \label{fig:ablation}
    \vspace*{-5pt}
\end{figure}
\mypar{Discrete scans across multiple directions.} Fig.\ref{fig:ablation}, \textit{left} demonstrates that accuracy improves consistently 
with the number of directional scans. With only \textbf{two directions}, spatial propagation remains limited, restricting how features interact across the image plane. Expanding to \textbf{four directions} strengthens vertical and horizontal flow, yielding a performance gain over two directions only (\textbf{+0.3\%}). Incorporating all \textbf{eight canonical orientations} further enhances isotropic information exchange by enabling diagonal and reverse traversal paths. Crucially, this improvement is not achieved through a naive raster scan over a single long sequence. Instead, our method performs discrete multi-line scans in each orientation, ensuring that state propagation resets appropriately and avoids the accumulation bias inherent to continuous 1-D ordering. Each directional update is governed by a normalized transition kernel, preventing instability and preserving balanced influence across directions. Finally, our directional traversal-selection mechanism ensures that aggregation across all eight paths remains coherent and complementary. Together, these design choices enable richer, more stable multi-directional propagation, leading to the highest Top-1 accuracy with eight directions.

\mypar{Effect of Traversal Selection Mechanism.}
We evaluate the impact of the Traversal Selection module by comparing Top-1 accuracy with and without this component Fig.~\ref{fig:ablation}, \emph{right}. The baseline model without Traversal Selection reaches \textbf{85.69\%} accuracy, while enabling the mechanism improves performance to \textbf{86.60\%}, yielding a clear gain of \textbf{+0.91\%}. This improvement highlights the benefit of selectively reweighting directional scans: instead of uniformly fusing all traversals, the module amplifies the most informative paths and suppresses less useful ones. As a result, feature aggregation becomes more discriminative, demonstrating that regulating directional influence is crucial for strengthening representation quality.

\begin{figure}[t]
    \centering
    \includegraphics[width=0.96\columnwidth]{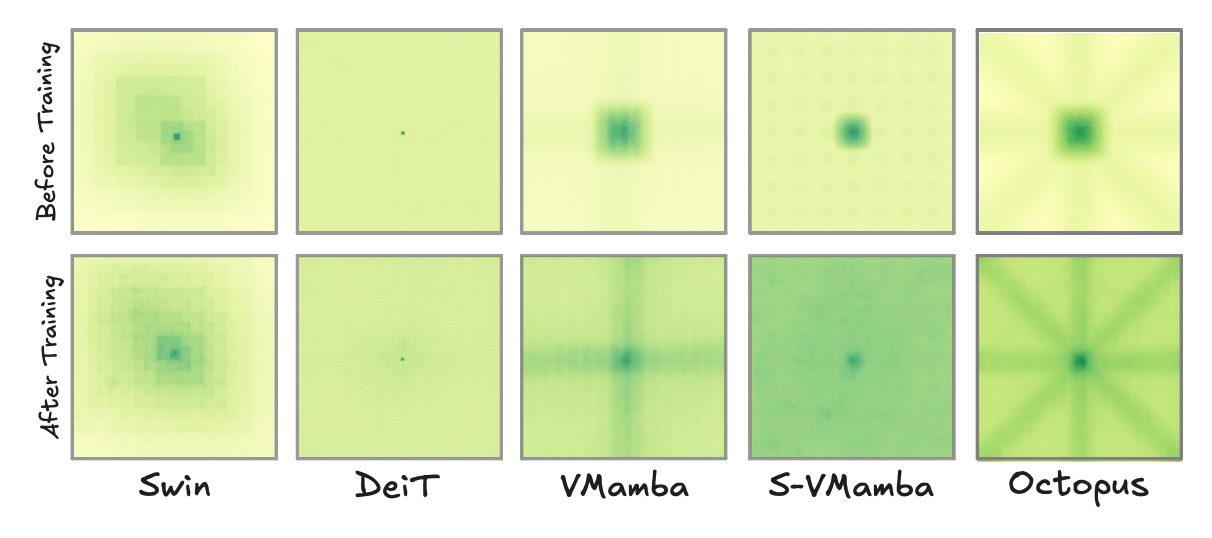}

    \vspace{-5pt}
    \caption{\textbf{ERF evolution across models} Comparison of effective receptive fields (ERFs) before and after training across different architectures. \ours{} shows the most structured and spatially aware ERF pattern.}
    \label{fig:erf}

    \vspace*{-10pt}
\end{figure}

\mypar{Spatial Isotropy Analysis.}
Figure~\ref{fig:erf} illustrates the evolution of the \emph{Effective Receptive Field} (ERF) before and after training across Swin, DeiT, VMamba, S-VMamba, and our \ours{} model. Before training, all architectures exhibit diffuse and weakly structured ERFs. After training, Swin retains window-induced block artifacts, DeiT forms smooth yet isotropic responses, and VMamba/S-VMamba develop anisotropic patterns caused by their raster-scan bias. In contrast, \emph{\ours{}} produces a highly structured and symmetric ERF with strong responses along eight canonical directions, indicating superior spatial connectivity and balanced directional propagation. This demonstrates \ours{}’s ability to learn a more spatially aware and geometrically consistent receptive field than prior SSM- and ViT-based models.

\begin{figure}[t]
    \centering
    \includegraphics[width=\columnwidth]{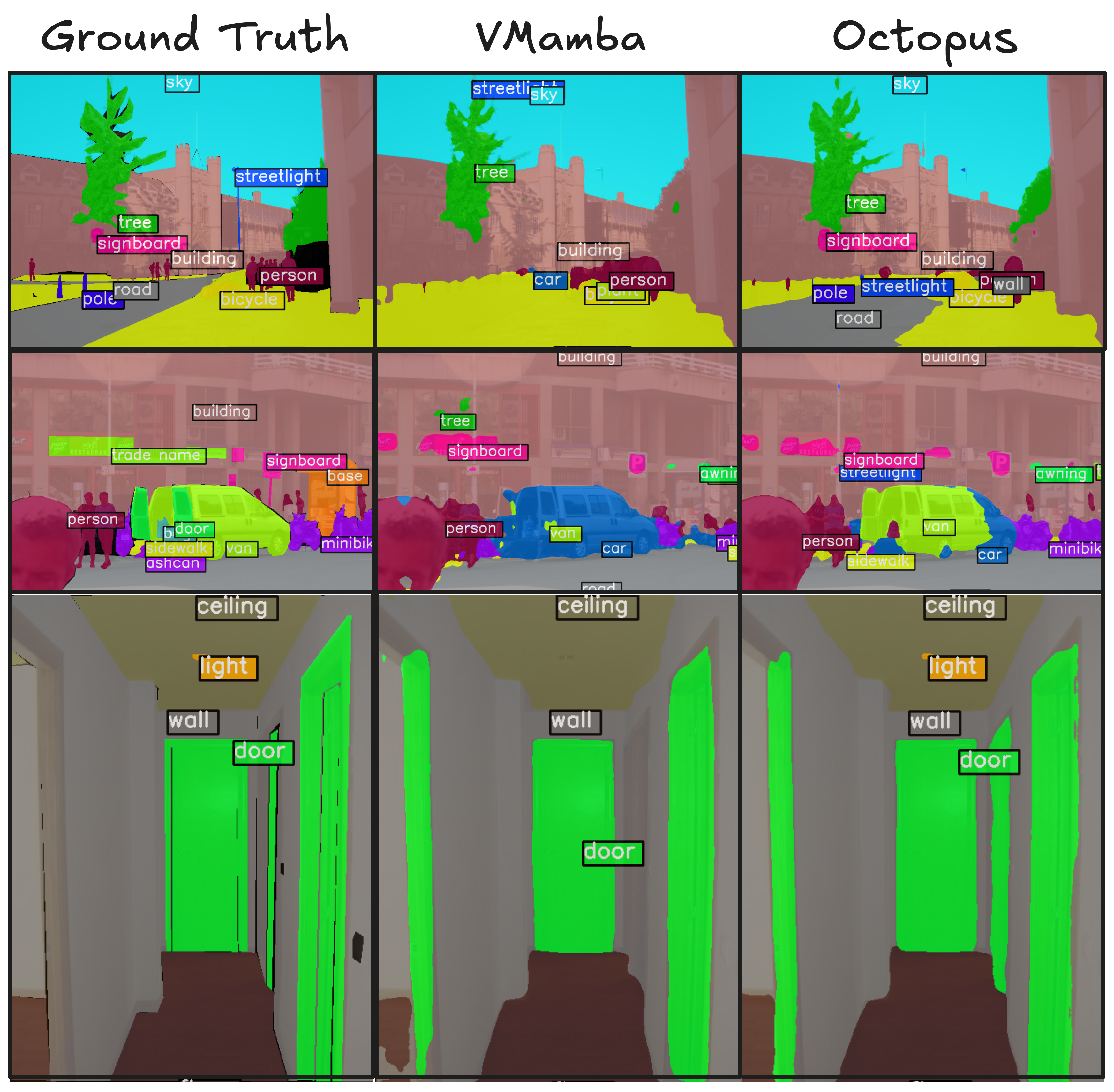}

    \vspace{-5pt}
    \caption{\textbf{Qualitative comparison of segmentation results.} \ours{} produces more complete and spatially consistent predictions than VMamba across diverse scenes. In the first example, \ours{} correctly segments the \textit{road} and yields a cleaner mask for the \textit{van}. In the indoor scene, \ours{} additionally identifies the \textit{ceiling light} and the \textit{side door}, which VMamba fails to detect. Overall, \ours{} demonstrates stronger boundary preservation and improved object completeness.}
    \label{fig:seg}

    \vspace*{-10pt}
\end{figure}

\mypar{Segmentation Map Comparison.} The qualitative comparison in Fig.~\ref{fig:seg} shows that \emph{\ours{}} produces more spatially coherent and complete segmentation compared to both the \emph{ground truth} and \emph{VMamba}. In the outdoor scene, \ours{} successfully recovers the \textit{road} region that VMamba partially misses, and it yields a noticeably cleaner and more accurate mask for the \textit{van}. In the crowded street example, \ours{} provides sharper object boundaries and reduces the fragmented predictions observed in VMamba. In the indoor corridor, \ours{} correctly identifies the \textit{ceiling light} and also segments the \textit{side door}, which VMamba fails to detect. Overall, these results demonstrate that \ours{} achieves stronger spatial reasoning, better object completeness, and improved robustness across diverse environments.

\section{Conclusion}
\label{sec:conclusion}

We introduced \ours{}, a spatially aware Vision State-Space Model that addresses the anisotropy of prior VSSMs. Instead of a single continuous raster scan, \ours{} performs \emph{multiple discrete scans} across eight canonical directions, enabling geometry-aligned state propagation that reduces directional bias and strengthens fine-grained spatial interactions. A normalized transition kernel and traversal-selection mechanism further ensure stable, spatially coherent multi-directional fusion.
Extensive experiments show that \ours{} delivers consistent improvements over VMamba and SpectralVMamba on both classification and segmentation. On ADE20K, it achieves notable gains and qualitatively recovers missing regions, sharpens boundaries, and segments small structure such as lights, all while preserving the linear-time efficiency of SSMs.

Overall, OCTOPUS shows that multi-directional \emph{discrete scanning} is a simple yet powerful way to inject strong spatial inductive bias into VSSMs. This opens the door to more spatially structured SSMs, richer geometric traversal designs, and hybrid SSM–attention models for robust and efficient visual understanding.

{
    \small
    \bibliographystyle{ieeenat_fullname}
    \bibliography{main}

@String(IJCV = {Int. J. Comput. Vis.})

@String(CVPR= {IEEE Conf. Comput. Vis. Pattern Recog.})

@String(ICCV= {Int. Conf. Comput. Vis.})

@String(NIPS= {Adv. Neural Inform. Process. Syst.})

@String(IJCV  = {IJCV})

@String(CVPR  = {CVPR})

@String(ICCV  = {ICCV})

@String(NIPS  = {NeurIPS})

@String(ICML = {Int. Conf. Machine Learning})

@article{vaswani2017attention,
  title={Attention is all you need},
  author={Vaswani, Ashish and Shazeer, Noam and Parmar, Niki and Uszkoreit, Jakob and Jones, Llion and Gomez, Aidan N and Kaiser, {\L}ukasz and Polosukhin, Illia},
  journal={Advances in neural information processing systems},
  volume={30},
  year={2017}
}

@article{gu2021efficiently,
  title={Efficiently modeling long sequences with structured state spaces},
  author={Gu, Albert and Goel, Karan and R{\'e}, Christopher},
  journal={arXiv preprint arXiv:2111.00396},
  year={2021}
}

@article{smith2022simplified,
  title={Simplified state space layers for sequence modeling},
  author={Smith, Jimmy TH and Warrington, Andrew and Linderman, Scott W},
  journal={arXiv preprint arXiv:2208.04933},
  year={2022}
}

@article{mamba,
  title={Mamba: Linear-Time Sequence Modeling with Selective State Spaces},
  author={Gu, Albert and Dao, Tri},
  journal={arXiv preprint arXiv:2312.00752},
  year={2023}
}

@inproceedings{mamba2,
  title={Transformers are {SSM}s: Generalized Models and Efficient Algorithms Through Structured State Space Duality},
  author={Dao, Tri and Gu, Albert},
  booktitle={International Conference on Machine Learning (ICML)},
  year={2024}
}

@article{huang2024localmamba,
  title={Localmamba: Visual state space model with windowed selective scan},
  author={Huang, Tao and Pei, Xiaohuan and You, Shan and Wang, Fei and Qian, Chen and Xu, Chang},
  journal={arXiv preprint arXiv:2403.09338},
  year={2024}
}

@article{xie2024quadmamba,
  title={QuadMamba: Learning Quadtree-based Selective Scan for Visual State Space Model},
  author={Xie, Fei and Zhang, Weijia and Wang, Zhongdao and Ma, Chao},
  journal={arXiv preprint arXiv:2410.06806},
  year={2024}
}

@misc{mahatha2025nerveneighbourhoodentropyguided,
      title={NERVE: Neighbourhood \& Entropy-guided Random-walk for training free open-Vocabulary sEgmentation}, 
      author={Kunal Mahatha and Jose Dolz and Christian Desrosiers},
      year={2025},
      eprint={2511.08248},
      archivePrefix={arXiv},
      primaryClass={cs.CV},
      url={https://arxiv.org/abs/2511.08248}, 
}

@inproceedings{hu2018squeeze,
  title={Squeeze-and-excitation networks},
  author={Hu, Jie and Shen, Li and Sun, Gang},
  booktitle={Proceedings of the IEEE conference on computer vision and pattern recognition},
  pages={7132--7141},
  year={2018}
}

@article{scanvmamba,
author = {Zhu, Qinfeng and Fang, Yuan and Cai, Yuanzhi and Cheng, Chen and Fan, Lei},
year = {2024},
month = {01},
pages = {1-14},
title = {Rethinking Scanning Strategies With Vision Mamba in Semantic Segmentation of Remote Sensing Imagery: An Experimental Study},
volume = {PP},
journal = {IEEE Journal of Selected Topics in Applied Earth Observations and Remote Sensing},
doi = {10.1109/JSTARS.2024.3472296}
}

@article{fu2022hungry,
  title={Hungry hungry hippos: Towards language modeling with state space models},
  author={Fu, Daniel Y and Dao, Tri and Saab, Khaled K and Thomas, Armin W and Rudra, Atri and R{\'e}, Christopher},
  journal={arXiv preprint arXiv:2212.14052},
  year={2022}
}

@article{liu2024vmamba,
  title={VMamba: Visual State Space Model},
  author={Liu, Yue and Tian, Yunjie and Zhao, Yuzhong and Yu, Hongtian and Xie, Lingxi and Wang, Yaowei and Ye, Qixiang and Liu, Yunfan},
  journal={arXiv preprint arXiv:2401.10166},
  year={2024}
}

@article{zhu2024vision,
  title={Vision mamba: Efficient visual representation learning with bidirectional state space model},
  author={Zhu, Lianghui and Liao, Bencheng and Zhang, Qian and Wang, Xinlong and Liu, Wenyu and Wang, Xinggang},
  journal={arXiv preprint arXiv:2401.09417},
  year={2024}
}

@article{shi2024multi,
  title={Multi-Scale VMamba: Hierarchy in Hierarchy Visual State Space Model},
  author={Shi, Yuheng and Dong, Minjing and Xu, Chang},
  journal={arXiv preprint arXiv:2405.14174},
  year={2024}
}

@article{gu2023mamba,
  title={Mamba: Linear-time sequence modeling with selective state spaces},
  author={Gu, Albert and Dao, Tri},
  journal={arXiv preprint arXiv:2312.00752},
  year={2023}
}

@article{hatamizadeh2024mambavision,
  title={Mambavision: A hybrid mamba-transformer vision backbone},
  author={Hatamizadeh, Ali and Kautz, Jan},
  journal={arXiv preprint arXiv:2407.08083},
  year={2024}
}

@article{gu2021combining,
  title={Combining recurrent, convolutional, and continuous-time models with linear state space layers},
  author={Gu, Albert and Johnson, Isys and Goel, Karan and Saab, Khaled and Dao, Tri and Rudra, Atri and R{\'e}, Christopher},
  journal={Advances in neural information processing systems},
  volume={34},
  pages={572--585},
  year={2021}
}

@InProceedings{Bahri_2025_CVPR,
    author    = {Bahri, Ali and Yazdanpanah, Moslem and Noori, Mehrdad and Dastani, Sahar and Cheraghalikhani, Milad and Hakim, Gustavo Adolfo Vargas and Osowiechi, David and Beizaee, Farzad and Ben Ayed, Ismail and Desrosiers, Christian},
    title     = {Spectral Informed Mamba for Robust Point Cloud Processing},
    booktitle = {Proceedings of the IEEE/CVF Conference on Computer Vision and Pattern Recognition (CVPR)},
    month     = {June},
    year      = {2025},
    pages     = {11799-11809}
}

@article{mehta2022long,
  title={Long range language modeling via gated state spaces},
  author={Mehta, Harsh and Gupta, Ankit and Cutkosky, Ashok and Neyshabur, Behnam},
  journal={arXiv preprint arXiv:2206.13947},
  year={2022}
}

@article{zhou2019semantic,
  title={Semantic understanding of scenes through the ade20k dataset},
  author={Zhou, Bolei and Zhao, Hang and Puig, Xavier and Xiao, Tete and Fidler, Sanja and Barriuso, Adela and Torralba, Antonio},
  journal={International Journal of Computer Vision},
  volume={127},
  pages={302--321},
  year={2019},
  publisher={Springer}
}

@article{ng2001spectral,
  title={On spectral clustering: Analysis and an algorithm},
  author={Ng, Andrew and Jordan, Michael and Weiss, Yair},
  journal={Advances in neural information processing systems},
  volume={14},
  year={2001}
}

@article{imagenet15russakovsky,
    Author = {Olga Russakovsky and Jia Deng and Hao Su and Jonathan Krause and Sanjeev Satheesh and Sean Ma and Zhiheng Huang and Andrej Karpathy and Aditya Khosla and Michael Bernstein and Alexander C. Berg and Li Fei-Fei},
    Title = { {ImageNet Large Scale Visual Recognition Challenge} },
    Year = {2015},
    journal   = {International Journal of Computer Vision (IJCV)},
    doi = {10.1007/s11263-015-0816-y},
    volume={115},
    number={3},
    pages={211-252}
}

@article{xcit,
	title        = {Xcit: Cross-covariance image transformers},
	author       = {Ali, Alaaeldin and Touvron, Hugo and Caron, Mathilde and Bojanowski, Piotr and Douze, Matthijs and Joulin, Armand and Laptev, Ivan and Neverova, Natalia and Synnaeve, Gabriel and Verbeek, Jakob and others},
	year         = 2021,
	journal      = NIPS,
	volume       = 34,
	pages        = {20014--20027}
}

@inproceedings{DeiT2021,
	title        = {Training data-efficient image transformers {\&} distillation through attention},
	author       = {Hugo Touvron and Matthieu Cord and Matthijs Douze and Francisco Massa and Alexandre Sablayrolles and Herv{\'{e}} J{\'{e}}gou},
	year         = 2021,
	booktitle    = ICML,
	pages        = {10347--10357}
}

@inproceedings{Swin2021,
	title        = {Swin Transformer: Hierarchical Vision Transformer using Shifted Windows},
	author       = {Liu, Ze and Lin, Yutong and Cao, Yue and Hu, Han and Wei, Yixuan and Zhang, Zheng and Lin, Stephen and Guo, Baining},
	year         = 2021,
	booktitle    = ICCV,
	pages        = {10012--10022}
}

@inproceedings{zhou2017scene,
	title        = {Scene parsing through ade20k dataset},
	author       = {Zhou, Bolei and Zhao, Hang and Puig, Xavier and Fidler, Sanja and Barriuso, Adela and Torralba, Antonio},
	year         = 2017,
	booktitle    = CVPR,
	pages        = {5122--5130}
}

@inproceedings{doersch2015unsupervised,
	title        = {Unsupervised visual representation learning by context prediction},
	author       = {Doersch, Carl and Gupta, Abhinav and Efros, Alexei A},
	year         = 2015,
	booktitle    = ICCV,
	pages        = {1422--1430}
}

@inproceedings{zhang20252dmamba,
  title={2DMamba: Efficient state space model for image representation with applications on giga-pixel whole slide image classification},
  author={Zhang, Jingwei and Nguyen, Anh Tien and Han, Xi and Trinh, Vincent Quoc-Huy and Qin, Hong and Samaras, Dimitris and Hosseini, Mahdi S},
  booktitle={Proceedings of the Computer Vision and Pattern Recognition Conference},
  pages={3583--3592},
  year={2025}
}

@inproceedings{baron20242,
  title={A 2-dimensional state space layer for spatial inductive bias},
  author={Baron, Ethan and Zimerman, Itamar and Wolf, Lior},
  booktitle={The Twelfth International Conference on Learning Representations},
  year={2024}
}

@inproceedings{fillioux2023structured,
  title={Structured state space models for multiple instance learning in digital pathology},
  author={Fillioux, Leo and Boyd, Joseph and Vakalopoulou, Maria and Courn{\`e}de, Paul-Henry and Christodoulidis, Stergios},
  booktitle={International Conference on Medical Image Computing and Computer-Assisted Intervention},
  pages={594--604},
  year={2023},
  organization={Springer}
}

@inproceedings{dastani2025spectral,
  title={Spectral State Space Model for Rotation-Invariant Visual Representation Learning},
  author={Dastani, Sahar and Bahri, Ali and Yazdanpanah, Moslem and Noori, Mehrdad and Osowiechi, David and Hakim, Gustavo Adolfo Vargas and Beizaee, Farzad and Cheraghalikhani, Milad and Mondal, Arnab Kumar and Lombaert, Herve and others},
  booktitle={Proceedings of the Computer Vision and Pattern Recognition Conference},
  pages={23881--23890},
  year={2025}
}
}

\clearpage
\setcounter{page}{1}
\setcounter{section}{0}
\renewcommand{\thesection}{\Alph{section}}
\renewcommand{\thesubsection}{\thesection.\arabic{subsection}}
\maketitlesupplementary

\begin{figure*}[ht!]
    \centering
    \includegraphics[width=1.0\textwidth]{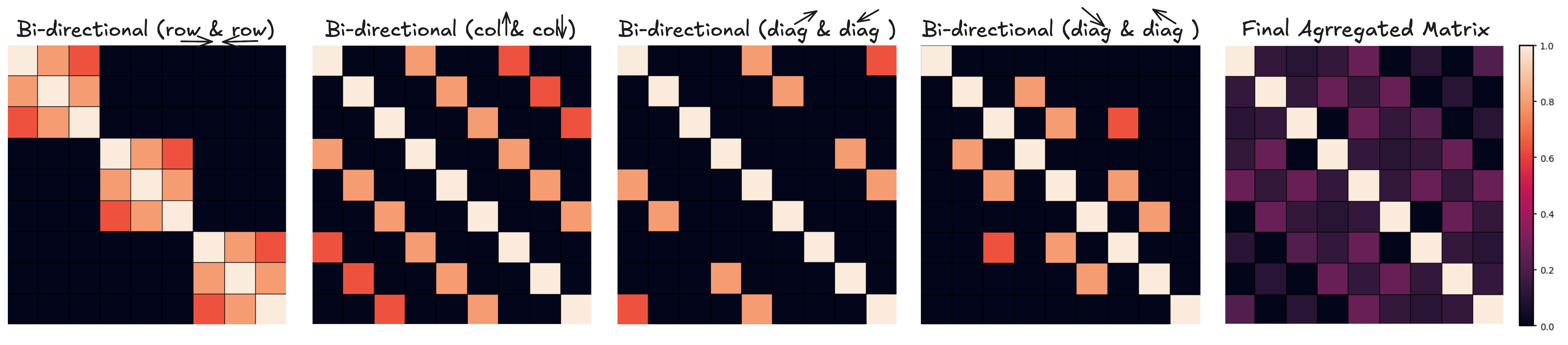}

    \vspace{-5pt}
    \caption{\textbf{Toy experiment illustrating Matrix $\mathbf{M}$.}
    Each bi-directional matrix encodes affinities along a single traversal direction (row, column, or diagonals). Aggregating them yields a more isotropic affinity structure that motivates the design of $\mathbf{M}$.}

    \label{fig:m-motivate}

    \vspace{-5pt}
\end{figure*}

\section{Discussion on Multi-Directional Scanning}

\mypar{Why simply increasing multiple raster-scans does not resolve recurrence bottlenecks in VMamba?}
Although recent VMamba-based architectures often attempt to enhance spatial modeling by extending the scanning process to four or even eight directions, prior analyses show that such multidirectional designs do not fundamentally resolve the limitations of sequential recurrence. As demonstrated in \emph{Rethinking Scanning Strategies With Vision Mamba in Semantic Segmentation of Remote Sensing Imagery: An Experimental Study}~\cite{scanvmamba}, the recurrence in Mamba continues to propagate information along a single serialized path, regardless of how many directional scans are introduced. Consequently, even diagonal, serpentine, or vertical traversal patterns behave as elongated one-dimensional sequences that fail to preserve true two-dimensional neighborhood structure. Their extensive evaluation of 22 scanning strategies further shows that increasing the number of scan directions yields negligible improvements, and may even degrade performance, since the underlying anisotropic modeling bias remains unchanged. These findings confirm that simply adding more scan directions does not address the mismatch between linear recurrence and the inherently two-dimensional geometry of dense prediction tasks.

\mypar{Motivation for the Construction of Matrix $\mathbf{M}$.}
Figure \ref{fig:m-motivate} presents a toy experiment that visualizes how each of the four bi-directional traversal patterns (row--row, column--column, and the two diagonal orientations) captures a different notion of long-range spatial affinity on the image grid. Each sub-matrix encodes a structured recurrence pattern: (i) the \emph{row bi-directional} matrix strengthens interactions between tokens lying on the same horizontal scan-line; (ii) the \emph{column bi-directional} matrix analogously reinforces vertical relationships; (iii--iv) the \emph{two diagonal} matrices capture affinities along the $\text{TopLeft}\!\to\!\text{BottomRight}$ and $\text{TopRight}\!\to\!\text{BottomLeft}$ diagonals, which are typically ignored in standard 1D scanning schemes used in state-space models.

Individually, each recurrence matrix is highly anisotropic: strong connectivity exists only along one geometric direction while other neighbors remain disconnected. By aggregating all four bi-directional
matrices, the toy experiment demonstrates how a unified affinity matrix $\mathbf{M}$ can approximate true $2$D neighborhoods. The final aggregated
matrix preserves directional strengths while providing isotropic coverage of local and semi-local spatial relationships. This motivates our use of $\mathbf{M}$ as a principled, geometry-aware propagation operator that reconciles the limitations of single-direction recurrences and enables more faithful spatial context modeling in vision state-space architectures.

\begin{algorithm}[H]
\caption{(O-Attention) Traversal Selection}
\label{algo:traversal_selection}
\begin{small}
\begin{algorithmic}[1]

\State \textbf{input:} feature tensor $\mathcal{X} \in \mathbb{R}^{B \times C \times D \times H \times W}$,
       learnable scoring network $f(\cdot)$
\State \textbf{output:} direction-weighted tensor 
       $\mathcal{Y} \in \mathbb{R}^{B \times C \times D \times H \times W}$

\Statex \textcolor{gray}{// Flatten spatial grid but keep directions.}
\State $\mathcal{X}_{\text{flat}} \gets \text{reshape}(\mathcal{X})$ 
       to $(B, C, D, HW)$

\Statex \textcolor{gray}{// Compute per-direction per-patch scores.}
\State $\mathcal{S} \gets f(\mathcal{X}_{\text{flat}})$            \hfill 
       \textcolor{gray}{$\mathcal{S} \in \mathbb{R}^{B \times D \times HW}$}

\Statex \textcolor{gray}{// Normalize scores across directions.}
\State $A_{\text{dir}} \gets \text{Softmax}(\mathcal{S}, \text{dim}=D)$
\State $A_{\text{dir}} \gets \text{unsqueeze}(A_{\text{dir}}, \text{dim}=2)$
       \hfill \textcolor{gray}{$(B,1,D,HW)$}

\Statex \textcolor{gray}{// Apply directional weights.}
\State $\mathcal{Y}_{\text{flat}} \gets \mathcal{X}_{\text{flat}} \odot A_{\text{dir}}$

\Statex \textcolor{gray}{// Restore spatial layout.}
\State $\mathcal{Y} \gets \text{reshape}(\mathcal{Y}_{\text{flat}})$ 
       to $(B, C, D, H, W)$

\State \Return $\mathcal{Y}$

\end{algorithmic}
\end{small}
\end{algorithm}

\begin{algorithm}[H]
\caption{Build Index Tensor}
\label{algo:index_tensors_simplified}
\begin{small}
\begin{algorithmic}[1]

\State \textbf{input:} grid height $H$, width $W$ 
\State \textbf{output:} index tensor $\texttt{idx}$, mask tensor $\texttt{mask}$, metadata $\texttt{meta}$

\Statex \textcolor{gray}{// Build a grid of linear indices.}
\State $\texttt{grid}[i,j] \gets i \cdot W + j$  \hfill \textcolor{gray}{($H \times W$)}

\Statex \textcolor{gray}{// 1. Generate directional scan-lines.}
\State $\mathcal{D} \gets$ set of traversal directions (e.g., rows, columns, diagonals)
\State $\texttt{seqs} \gets$ list of scan-line sequences extracted from $\texttt{grid}$ for each $d \in \mathcal{D}$

\Statex \textcolor{gray}{// 2. Compute padding requirements.}
\State $n_{\max} \gets \max\limits_{d \in \mathcal{D}} |\texttt{seqs}[d]|$
\State $L_{\max} \gets \max\limits_{\ell \in \texttt{seqs}[d]} |\ell|$

\Statex \textcolor{gray}{// 3. Allocate padded tensors.}
\State $\texttt{idx}  \gets$ tensor of shape $(|\mathcal{D}|, n_{\max}, L_{\max})$ filled with $-1$
\State $\texttt{mask} \gets$ boolean tensor of same shape, initialized to \texttt{False}

\Statex \textcolor{gray}{// 4. Write valid scan-line indices.}
\For{each direction $d \in \mathcal{D}$}
    \For{each line index $t$ and sequence $\ell$ in $\texttt{seqs}[d]$}
        \State Insert elements of $\ell$ into $\texttt{idx}[d, t, :|\ell|]$
        \State Set $\texttt{mask}[d, t, :|\ell|] \gets \texttt{True}$
    \EndFor
\EndFor

\Statex \textcolor{gray}{// 5. Metadata.}
\State $\texttt{meta} \gets \{H, W, n_{\max}, L_{\max}\}$

\State \Return $\texttt{idx},\,\texttt{mask},\,\texttt{meta}$

\end{algorithmic}
\end{small}
\end{algorithm}

\begin{algorithm}[H]
\caption{O-Scan}
\label{algo:o-scan-generic}
\begin{small}
\begin{algorithmic}[1]

\State \textbf{input:} feature map $x \in \mathbb{R}^{B \times C \times H \times W}$
\State \textbf{output:} directional scan tensor $x_s$, scan mask $m_s$, metadata $\texttt{meta}$

\Statex \textcolor{gray}{// 1. Obtain scan-line definitions.}
\State $(\texttt{idx},\, \texttt{mask},\, \texttt{meta}) \gets$ precomputed directional scan-lines 
      for grid size $(H,W)$
      \hfill \textcolor{gray}{indices and validity masks}

\Statex \textcolor{gray}{// 2. Flatten spatial grid.}
\State $x_{\text{flat}} \gets$ reshape $x$ to $(B, C, HW)$

\Statex \textcolor{gray}{// 3. Extract directional sequences.}
\State $x_s \gets$ gather elements of $x_{\text{flat}}$ along indices in $\texttt{idx}$
      \hfill \textcolor{gray}{produces $(B, C, D, n_{\text{lines}}, L_{\max})$}

\Statex \textcolor{gray}{// 4. Apply scan-line masks.}
\State $m_s \gets$ broadcast $\texttt{mask}$ to match the shape of $x_s$
\State $x_s \gets x_s \odot m_s$ 

\Statex \textcolor{gray}{// 5. Return directional sequences and metadata.}
\State augment $\texttt{meta}$ with batch size, channels, and direction count
\State \Return $(x_s,\, m_s,\, \texttt{meta})$
\end{algorithmic}
\end{small}
\end{algorithm}

\begin{algorithm}[H]
\caption{O-Merge}
\label{algo:o-merge-generic}
\begin{small}
\begin{algorithmic}[1]

\State \textbf{input:} directional sequences $x_s$, metadata $\texttt{meta}$
\State \textbf{output:} reconstructed feature map 
      $Y \in \mathbb{R}^{B \times C \times H \times W}$

\Statex \textcolor{gray}{// 1. Read metadata and reshape sequences.}
\State Extract $B,\, C,\, D,\, n_{\text{lines}},\, L_{\max},\, H,\, W$ from $\texttt{meta}$
\State Reshape $x_s$ to $(B,\, C,\, D,\, n_{\text{lines}},\, L_{\max})$

\Statex \textcolor{gray}{// 2. Apply directional fusion (O-Attention).}
\State $x_f \gets \textsc{O\text{-}Attention}(x_s)$
      \hfill \textcolor{gray}{direction-weighted sequences}

\Statex \textcolor{gray}{// 3. Mask out invalid scan positions.}
\State Retrieve index tensor $\texttt{idx}$ and mask $\texttt{mask}$ from $\texttt{meta}$
\State $x_f \gets x_f \odot \texttt{mask}$

\Statex \textcolor{gray}{// 4. Scatter fused values back to the image grid.}
\State Initialize accumulation buffer 
      $\texttt{buf} \in \mathbb{R}^{B \times C \times HW}$ as zeros
\State For each direction and scan-line, place values of $x_f$ at locations given by $\texttt{idx}$
      \hfill \textcolor{gray}{overlapping contributions add together}

\Statex \textcolor{gray}{// 5. Restore 2D spatial layout.}
\State $Y \gets \text{reshape}(\texttt{buf})$ to $(B,\, C,\, H,\, W)$
\State \Return $Y$
\end{algorithmic}
\end{small}
\end{algorithm}

\section{Pseudocode and Implementation Notes }

\begin{figure*}[t!]
    \centering
    \includegraphics[width=.9\textwidth]{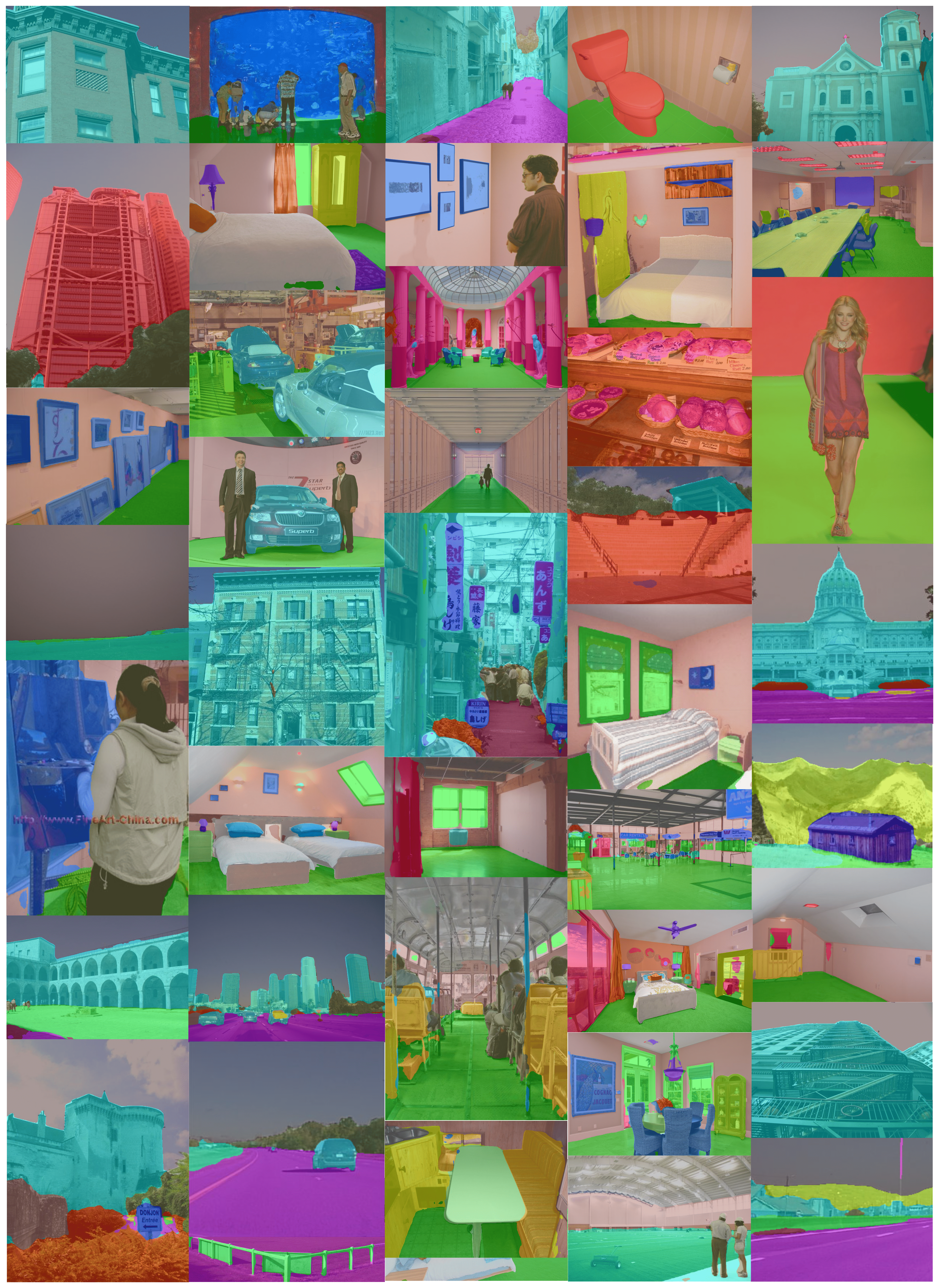}

    \vspace{-5pt}
    \caption{\textbf{Additional segmentation results.}}

    \label{fig:seg-final}

    \vspace{-5pt}
\end{figure*}

In this section, we provide the pseudo-code for the four modules that constitute our discrete multi-directional scanning
framework. These algorithms offer a concise, implementation-oriented
summary of the core components described in the main paper, and are
intended to complement the conceptual exposition provided there. They
serve both as a high-level abstraction of the underlying operations and
as a practical reference for reproducing our results.

We begin with Algorithm~\ref{algo:index_tensors_simplified}, which defines the \textit{BuildIndexTensors} module responsible for constructing the directional traversal structure of the image grid. As explained in the main text, this module enumerates scan-lines along all eight canonical directions and organizes them into padded index and mask tensors for efficient reuse. Algorithm~\ref{algo:o-scan-generic} then introduces the \textit{O-Scan} module,
which extracts directional sequences from an input feature map by gathering pixel values along the precomputed scan-lines. This step transforms the 2D grid into eight sets of structured 1D directional
representations. Next, Algorithm~\ref{algo:traversal_selection} describes the \textit{O-Attention} module. This component applies a lightweight scoring network to each directional sequence and computes a softmax across directions, yielding traversal-selection weights that determine how strongly each orientation should contribute at every spatial location. Finally, Algorithm~\ref{algo:o-merge-generic} outlines the \textit{O-Merge} module, which reconstructs the spatial feature map by projecting the direction-weighted sequences back to their original coordinates and aggregating overlapping contributions.

Together, these four algorithms provide a complete implementation of our discrete scanning pipeline and act as a practical guide for integrating multi-directional reasoning into dense prediction architectures.

\section{Additional Qualitative Results}

\begin{figure*}[t!]
    \centering
    \includegraphics[width=1.0\textwidth]{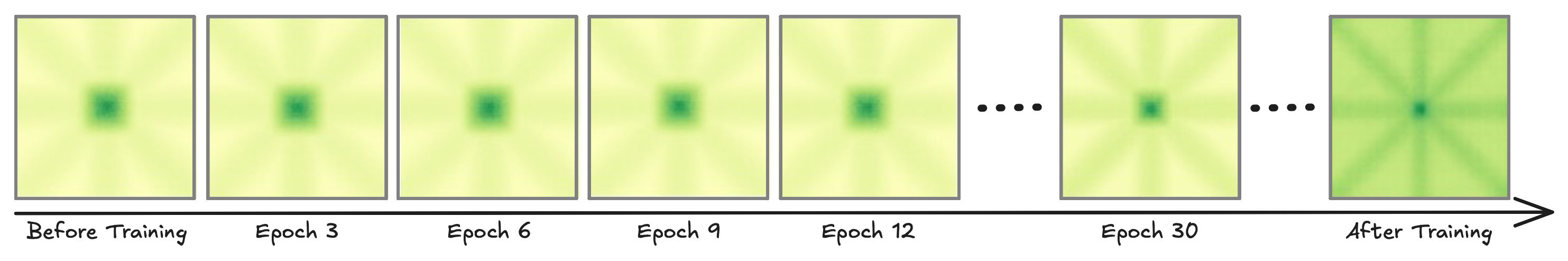}

    \vspace{-5pt}
    \caption{\textbf{Evolution of the Effective Receptive Field (ERF) during training.}  
    We show ERF visualizations across training epochs, illustrating how the receptive field patterns change from early to late stages of optimization.}

    \label{fig:erf-process}

    \vspace{-5pt}
\end{figure*}

\mypar{Discusion of the Evolution of the ERF Maps}
Figure~\ref{fig:erf-process} illustrates how we track the Effective Receptive Field (ERF) throughout training. For each selected epoch, we compute the ERF of the same reference token and visualize it as a heatmap, enabling a direct comparison of how the spatial influence pattern evolves from the randomly initialized model (“Before Training”) to intermediate checkpoints (Epochs 3–30) and finally to the fully trained model (“After Training”). From the ERF evolution, we observe that the model gradually transitions from an almost uniform and weakly structured receptive field to a highly organized, directional one. Early in training, the ERF is diffused and it lacks strong spatial preference, indicating that the model has not yet formed meaningful propagation patterns. As training process evolves , distinct directional spokes appear, showing that the model begins to exploit structured, anisotropic interactions across the image grid. By the end of the training, the ERF becomes sharper and more concentrated, demonstrating that the model has learned stable, spatially aware propagation pathways that align with its eight-directional scanning design.

\end{document}